\documentclass[procedia]{easychair}

\usepackage{doc}
\usepackage{makeidx}
\usepackage[linesnumbered,lined,boxed,commentsnumbered, ruled, vlined, noend]{algorithm2e}
\usepackage{floatrow}
\newfloatcommand{capbtabbox}{table}[][\FBwidth]
\usepackage{blindtext}
\usepackage[export]{adjustbox}

\def\procediaConference{99th Conference on Topics of
  Superb Significance (COOL 2014)}

\title{GPU-based pedestrian detection for autonomous driving}

\titlerunning{GPU-based pedestrian detection for autonomous driving}

\author{
    V. Campmany\inst{1,2}
\and
	S. Silva\inst{1,2}
\and
    A. Espinosa\inst{1}
\and
    J.C. Moure\inst{1}
\\
\and
    D. V{\'a}zquez\inst{2}
\and
    A. M. L{\'o}pez\inst{2}
}

\institute{
   Universitat Autonoma de Barcelona,
   Spain.\\ 
\and
   Computer Vision Centre (CVC),
   Spain.\\
}
   
\authorrunning{Victor Campmany et. al.}

\begin{document}
\maketitle
\vspace{-0.5cm}

\begin{abstract}
We propose a real-time pedestrian detection system for the embedded Nvidia Tegra X1 GPU-CPU hybrid platform. The pipeline is composed by the following state-of-the-art algorithms: Histogram of Local Binary Patterns (LBP) and Histograms of Oriented Gradients (HOG) features extracted from the input image; Pyramidal Sliding Window technique for candidate generation; and Support Vector Machine (SVM) for classification. Results show a 8x speedup in the target Tegra X1 platform and a better $performance/watt$ ratio than desktop CUDA platforms in study.

\end{abstract}


%
%

\vspace{-0.5cm}
\section{Introduction}
\vspace{-0.3cm}
Autonomous driving will improve safety, reduce pollution and congestion, and enable handicapped people, elderly persons and kids to have more freedom. It requires perceiving and understanding the vehicle environment (road, traffic signs, pedestrians, vehicles ...) using sensors (cameras, LIDAR’s, sonars, and radar), providing self-localization (with GPS, inertial sensors and visual localization in precise maps), controlling the vehicle, and planning the routes. The underlying algorithms are high computationally demanding and require real-time response.

A pedestrian detector that locates humans on a digital image is a key module for robotics application and autonomous vehicles. The wide variation by which humans appear with different poses, clothes, illuminations and backgrounds makes the problem one of the hardest within the computer vision field, and the source of an active research during the last twenty years ~\cite{geronimo-survey, gavrila-survey, caltech-survey, local-experts}. 

The real-time constraints of most applications of pedestrian detection are tight, and currently not attainable by general-purpose processors. Attaching a desktop GPU for performance acceleration is expensive in terms of space and power consumption. The recent appearance of embedded GPU-accelerated systems based on the Nvidia's Tegra X1 ARM processor, like the Jetson TX1 and DrivePX platforms, represents a promising approach for low-cost and low-consumption real-time pedestrian detection. 

We have designed a complete GPU-accelerated pedestrian detection pipeline\footnote{\url{https://github.com/vcampmany/CudaVisionSysDeploy}} based on ~\cite{local-experts}. We have analyzed the alternative parallelization schemes and data layouts of the underlying algorithms and selected general and scalable solutions that do not sacrifice performance nor detection accuracy. We have evaluated three pedestrian detection versions on desktop and embedded platforms and proved that: (1) real-time (20 images of 1242$\times$375 pixels per second) can be reached on an embedded GPU-accelerated system with state-of-the-art accuracy; (2) GPU-acceleration provides between 8x and 60x performance speedup with respect to a baseline multi-core CPU implementation; and (3) the Tegra X1 processor at least doubles the performance per watt of the system accelerated by a GTX 960 GPU.

The rest of the paper is organized as follows. Section \ref{sota} introduces the related work. Section \ref{hoglbp} describes the baseline pedestrian detector. In section \ref{dev} we analyze each algorithm and discuss different parallelization schemes. Finally, section \ref{experiments} provides the obtained results and section \ref{conclusions} summarizes the work.  


\vspace{-0.6cm}
\section{Related work} \label{sota}
\vspace{-0.3cm}

Computer Unified Device Architecture (CUDA) is a platform created by Nvidia to develop general purpose applications for the GPUs. Nvidia GPUs are composed by tens of processing units called \emph{Streaming Multiprocessor} (SMs). The SMs share a L2 cache and an external global memory. Each SM has a shared memory that is managed explicitly and a L1 cache. A CUDA \emph{kernel} is composed by thousands of threads executing the same program with different data. The threads are divided into groups of up to 1024 threads called Cooperative Thread Arrays (CTAs). The threads in a CTA collaborate using the on-chip shared memory. Each CTA is divided into batches of 32 threads called \emph{warps}. Finally, individual threads have a reserved memory region in each layer of the memory hierarchy called local memory.

A critical performance issue is the memory access pattern of the algorithm. GPUs achieve full memory performance when the memory accesses are \emph{coalesced}. Coalesced memory access refers to combining multiple memory operations into a single memory transactions. To achieve coalescing, the 32 threads of the warp must access consecutive memory addresses. The previously described constrains data layout, memory transfers and work distribution become key factors in order to achieve the best performance.

Since the appearance of GPGPU computational platforms, several object detection algorithms have been ported to the GPU. There are also examples of using Field Programmable Gate Array (FPGA) designs, obtaining outstanding results ~\cite{fpga-hog}. In comparison, the reduced development costs of the CUDA programming environment and the affordable desktop CUDA enabled GPU cards make them more suitable for testing new algorithms.

Works such as ~\cite{ped-100-fps} assert that exploiting the massively parallel paradigm for object detection algorithms outperforms a highly tuned CPU version ~\cite{fastest-west}. Previously related researches like ~\cite{sliding-win-gpu-wojek, sliding-win-gpu-zhang, hog-svm-gpu} developed a GPU object detector using the well-known HOG-SVM approach obtaining a performance boost. 

In the previous work, the evaluations use desktop GPUs to evaluate the algorithms proposed as a first step to evaluate the GPU as a suitable target platform. In this work, we propose a real-time pedestrian detector running on a low-consumption GPU devices like the Tegra X1 platform. We also present for the first time a GPU implementation of the HOGLBP-SVM detection pipeline ~\cite{hoglbp-wang}.
\vspace{-0.6cm}
\section{Pedestrian detection} \label{hoglbp}
\vspace{-0.3cm}


Any computer vision-based pedestrian detector is mainly composed by four core modules: the candidate generation, the feature extraction, the classification and the refinement. The candidate generation provides rectangular image windows which eventually contain pedestrians. These windows are described using distinctive patches in the feature extraction stage. During the classification stage the windows are labeled using a learned model accordingly to its features. Finally, as a pedestrian could be detected by several windows, they are finally managed in the refinement stage.

For the candidate generation we use the Pyramidal Sliding Window (SW) method. To extract the features of the images we use Local Binary Patterns (LBP) and Histograms of Oriented Gradients (HOG). A Support Vector Machine (SVM) is taken as classifier, and, finally the refinement is performed by the Non-maximum Suppression algorithm ~\cite{nms-laptev}.  
\vspace{-0.5cm}
\section{Design and Analysis of Massively-Parallel Algorithms} \label{dev}
\vspace{-0.3cm}
We have implemented three different detection pipelines combining the basic algorithms mentioned in section \ref{hoglbp}
. In this section we present the algorithms and the decisions behind their massively-parallel implementations on a CUDA architecture. We start describing the general detection pipelines and the design methodology, and then delve into the details of each algorithm. 

\vspace{-0.5cm}
\subsection{Overview of the Detection Pipelines}
\vspace{-0.3cm}
The three detection pipelines considered in this work, ordered from lower to higher accuracy and computational complexity, are LBP-SVM, HOG-SVM and HOGLBP-SVM. They represent three realistic options for an actual detection system, where one of them has to trade off functionality with processing rate. The pipelines follow the same scheme; differences appear on the feature extraction stage: LBP-SVM uses LBP, while HOG-SVM takes advantage of HOG. The HOGLBP-SVM pipeline combines both algorithms. As shown in previous research ~\cite{hoglbp-wang}, concatenating HOG and LBP feature vectors gives an significant increase of performance. 
The hybrid processing pipeline is designed to use the CPU (Host) and the GPU (Device): (1) the captured images are copied from the Host memory space to the Device; (2) the scaled-pyramid of images is created; (3) features are extracted from each pyramid layer; (4) every layer is segmented into windows to be classified; (5) detection results are copied to the CPU memory to execute the Non-maximum Suppression algorithm that refines the results.

\vspace{-0.45cm}
\subsection{Histograms of Local Binary Patterns (LBP)}
\vspace{-0.3cm}
LBP is a feature extraction method that gives information of the texture on a chunk of the image. The process can be divided into two steps: the \emph{LBP Map} computation and the \emph{LBP Histograms} computation.

The LBP Map ~\cite{lbp-ojala} is computationally classified as a 2-dimensional Stencil pattern algorithm. The central pixel is compared with each of its nearest neighbors; if the value is lower than the center a 0 is stored, otherwise, a 1 is stored.  Then, this binary code is converted to decimal to generate the output pixel value. 

\vspace{-0.1cm}
\begin{algorithm}[h] \label{lbp-pseudo}
 \SetAlgoLined
 \SetKwInOut{Input}{input}\SetKwInOut{Output}{output}
 
 \Input{I[H][W]}
 \Output{LBP[H][W]}
 \BlankLine
 parallel \For {y=0 \KwTo H and x=0 \KwTo W} {
   LBP[y][x] = \emph{LBPf(y, x)};
 }
 \caption{Massively parallel computation of the LBP map} 
\end{algorithm}
\vspace{-0.3cm}

\begin{figure}
\centering
	\includegraphics[width=\textwidth]{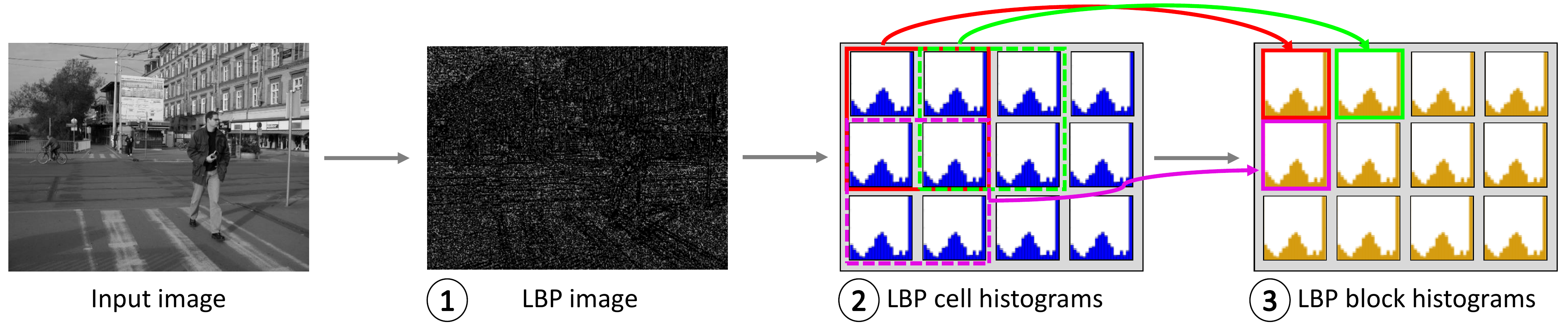}
  	\vspace{-0.3cm}
	\caption{LBP: (1) compute LBP value for each pixel (Algorithm \ref{lbp-pseudo}), (2) compute $8\times8$ pixels Cell Histograms (Algorithm \ref{cell-pseudo}), (3) compute $16\times16$ pixels block histograms (Algorithm \ref{block-pseudo}). }
	\label{lbp-pipe}
\end{figure}

Finally, we extract the image features by computing the LBP Histograms. Histograms of blocks of $16\times16$ pixels are computed over the LBP image. The histograms have a $50\%$ overlap in the $X$ and $Y$ axis meaning that each region will be redundantly computed 4 times. We avoided the redundant computing of the overlapped descriptors by splitting the \emph{Block Histograms} into smaller \emph{Cell Histograms} of $8\times8$ pixels. Then, these partial histograms are reduced in groups of four (\emph{histogram reduction}) to generate the output block histograms. Figure \ref{lbp-pipe} shows the previously described sequence of operations to compute the LBP.
\vspace{-0.3cm}

\vspace{-0.2cm}
\subsubsection{Analysis of parallelism \& CUDA mapping for computing LBP}
\vspace{-0.2cm}
We implemented the 2-dimensional Stencil pattern by mapping each thread to one output pixel (Algorithm \ref{lbp-pseudo}). With this work distribution there are no data dependences among threads. Each thread performs 9 \textit{reads} (the central value and the eight neighbors) and 1 \textit{store} and, the memory accesses are coalesced.

We designed two different solutions to compute the LBP histograms; the first, is a straightforward parallelization without thread collaboration (\emph{Na\"ive scheme}); the second, with thread collaboration, is designed to be more scalable (\emph{Scalable scheme}).

The Na\"{i}ve scheme follows a Map pattern: each thread generates a Cell Histogram. The histogram reduction is performed in the same way: each thread is mapped to a Block Histogram and the thread performs the histogram reduction.

The Scalable scheme solution used by our system aims for an efficient memory access and data reutilization. Each thread is mapped to an input pixel of the image, and using atomic operations each thread adds to its corresponding Cell Histogram (Scatter pattern, see Algorithm \ref{cell-pseudo}). To generate the Block Histograms every histogram reduction is performed by a warp (see Algorithm \ref{block-pseudo}). With this design we attain coalesced memory access which leads to an scalable algorithm for different image sizes. 

\vspace{-0.3cm}
\begin{algorithm}[ht] \label{cell-pseudo}
 \SetAlgoLined
 \SetKwInOut{Input}{input}\SetKwInOut{Output}{output}
 
 \Input{LBP[H][W]}
 \Output{CH[H/8][W/8][S]}
 \BlankLine
 parallel \For {y=0 \KwTo H and x=0 \KwTo W} {
   bin = LBP[y][x]\;   
   \emph{atomicAdd}(CH[y/8][x/8][bin], 1) \;
 }
 \caption{Scalable computation scheme of the Cell Histograms. Each thread reads a pixel and updates the corresponding cell. We use atomic operations (Read-Add-Store) to avoid data races. $S \Leftarrow histogram$.}
\end{algorithm}
\vspace{-0.3cm}

\begin{algorithm}[h] \label{block-pseudo}
 \SetAlgoLined
 \SetKwInOut{Input}{input}\SetKwInOut{Output}{output}
 
 \Input{CH[H/8][W/8][S]}
 \Output{Fa[Hb][Wb][S]}
 \BlankLine
 parallel \For {y=0 \KwTo Hb and x=0 \KwTo Wb}{
   SIMD parallel \For {lane=0 \KwTo WarpSize}{
   	 t = lane\;
     \While {$t \textless S$} {
     Fa[y][x][t] = \emph{hReduction}(t)\;
     t = t + \emph{WarpSize};
    }
  }   
 }
 \caption{Scalable computation of the histogram reduction to generate the Block Histograms (\emph{Fa}) \emph{hReduction}: $Hb\Leftarrow H/8-1$; $Wb \Leftarrow W/8-1$.}
\end{algorithm}
\subsection{Histogram of Oriented Gradients (HOG)}
\vspace{-0.3cm}
The method of Histograms of Oriented Gradients ~\cite{hog-dalal} counts the occurrence of gradient orientation on a chunk of the image. The process could be divided into two steps: \emph{Gradient} computation and the \emph{Histograms} computation.

Gradient computation is used to measure the directional change of color in an image. The algorithm follows a 2 dimensional Stencil pattern. The gradient of a pixel has two components, the magnitude ($\omega$) and the orientation ($\theta$). The orientation is the directional change of color and the magnitude gives us information of the intensity of the change. 

\begin{figure}[t]
\centering
	\includegraphics[width=\textwidth]{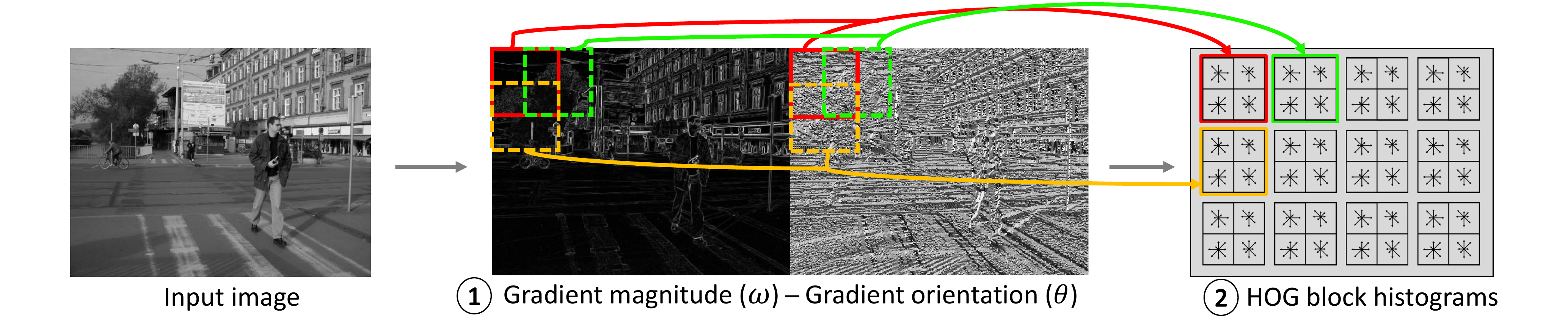}
	\caption{Steps to compute the HOG features: (1) given a grayscale image, compute the gradient (Algorithm \ref{gradient-pseudo}); (2) compute the Block Histograms with trilinear interpolation (Algorithm \ref{hog-pseudo}).}
	\label{hog-pipe}
\end{figure}

Histograms are computed by splitting the Gradient image into blocks of $16\times16$ pixels with $50\%$ overlap in X and Y axis (the same configuration as the LBP Histograms). In this case, because of histograms trilinear interpolation we can not compute $8\times8$ pixels Cell Histograms and then carry out the histogram reduction. Trilinear interpolation is used to avoid sudden changes in the Block Histograms vector (aliasing effect) ~\cite{hog-dalal}. Each Block Histogram is composed by four concatenated $8\times8$ pixels Cell Histograms. Different bins of the Block Histogram receive a weighted value of the orientation ($\theta$) multiplied by the magnitude of the gradient ($\omega$). Depending on the pixel coordinates, each input value could be binned into two, four or eight bins of the Block Histogram. The sequence of steps to compute the HOG features is described in Figure \ref{hog-pipe}.
\vspace{-0.7cm}
\subsubsection{Analysis of parallelism \& CUDA mapping for computing HOG}
\vspace{-0.2cm}
The gradient computation kernel follows a Stencil pattern: individual threads are mapped to each output pixel (Algorithm \ref{gradient-pseudo}). Single threads perform 4 \textit{reads} and 1 \textit{store}, and coalesced memory accesses are achieved.

The Histograms computation has been parallelized assigning each thread to the computation of one histogram (Large-grain task parallelism, see Algorithm \ref{hog-pseudo}). With this structure there is no collaboration among threads and memory accesses are not coalesced, though, the mapping avoids the use of the costly atomic memory operations.

We implemented three different kernels following the scheme in Algorithm \ref{hog-pseudo}. The first stores the data in global memory (\emph{HOG Global}). To reduce the latency of the global memory we designed two more kernels: one stores the histograms in local memory, taking advantage of the L1 cache (\emph{HOG Local}) and the other uses the on-chip shared memory (\emph{HOG Shared}). In section \ref{algo-res} we will discuss the results of the implementations.
\begin{algorithm} \label{gradient-pseudo}
 \SetAlgoLined
 \SetKwInOut{Input}{input}\SetKwInOut{Output}{output}
 
 \Input{I[H][W]}
 \Output{M[H][W], O[H][W]}
 \BlankLine
 parallel \For {y=0 \KwTo H and x=0 \KwTo W} {
   dx = I[y][x-1] - I[y][x+1]\;
   dy = I[y-1][x] - I[y+1][x]\;
   M[y][x] = \emph{sqrt}(dx $\ast$ dx, dy $\ast$ dy)\;
   O[y][x] = \emph{arctan}(dx, dy)\;
 }
 \caption{Massively parallel computation of the Gradient}
\end{algorithm} 

\begin{algorithm} \label{hog-pseudo}
 \SetAlgoLined
 \SetKwInOut{Input}{input}\SetKwInOut{Output}{output}
 
 \Input{M[H][W], O[H][W]}
 \Output{Fb[Hb][Wb][S]}
 \BlankLine
 parallel \For {y=0 \KwTo Hb and x=0 \KwTo Wb} {
   \For{i=0 to 16}{
     \For{j=0 to 16}{
       \emph{updateBlockHistogram}(Fb[y][x], i, j)\;
     }
   }
 }
 \caption{Massively parallel computation of the HOG Histograms. 
 }
\end{algorithm}

\vspace{-0.5cm}
\subsection{Pyramidal Sliding Window \& Support Vector Machine}
\vspace{-0.2cm}
Pyramidal Sliding Window creates multiple downscaled copies of the input images to detect pedestrians of various sizes and at different distances. Then, every copy is split into highly overlapped regions of $128\times 64$ pixels, called windows. Each window is described with a feature vector ($\vec{x}$). The vector is composed by the concatenation of the Block Histograms (computed with HOG and LBP) enclosed in the given region. Then, every vector is evaluated to predict if the region contains a pedestrian or not.

\begin{figure}[t]
\centering
	\includegraphics[width=\textwidth]{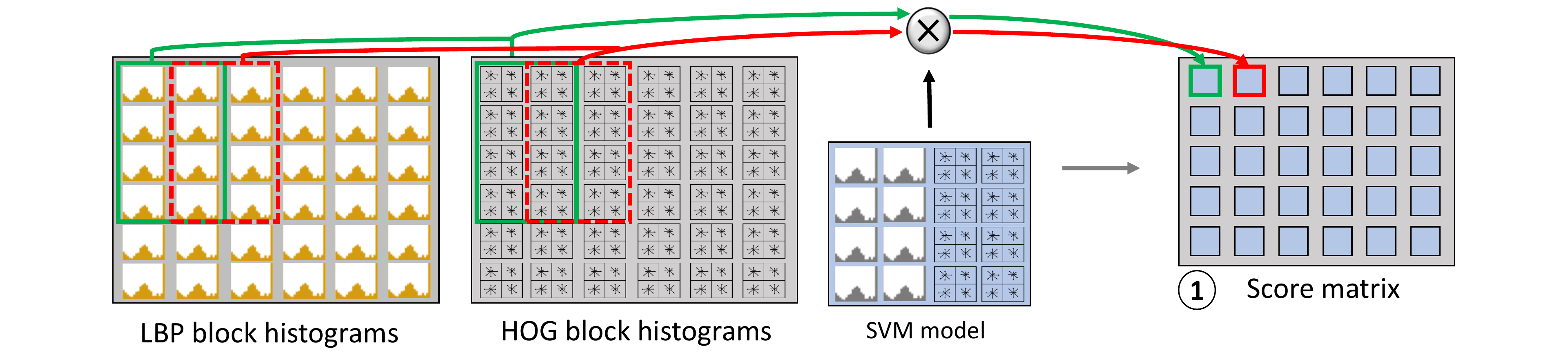}
	\caption{(1) Sliding Window and SVM inference of the HOG and LBP features. Each window is evaluated with Algorithm \ref{svm-pseudo}  computing the distance of the HOG+LBP features from the SVM}
	\label{svm-pipe}
\end{figure}

Support Vector Machine (SVM) is a supervised learning method that is able to discriminate two categories, in our case pedestrians from background ~\cite{svm-vapnik}. After training the SVM, we obtain a model that performs as an n-dimensional plane that distinguishes pedestrians from background. The SVM training is done offline. However, the SVM inference is done online. SVM gets as input a feature vector ($\vec{x}$) and computes its distance to the model hyper-plane ($\vec{\omega}$). This distance is computed with the dot product operation. Then, the window is classified as pedestrian if the distance is greater than a given threshold and as background otherwise. Figure \ref{svm-pipe} shows the steps needed to evaluate each window taking HOG and LBP features as the image descriptors.

\vspace{-0.5cm}
\subsubsection{Analysis of parallelism \& CUDA mapping for SW-SVM}
\vspace{-0.3cm}
We first implemented a na\"ive version (\emph{Na\"ive SVM}) with a Large-grain task parallelism and no thread collaboration. Each thread is responsible of the computation of the dot product between the candidate window ($\vec{x}$) and the model ($\vec{\omega}$). The approach was not scalable and became critical as it is the kernel with the largest workload of the pipeline.

To efficiently compute the dot product we designed a CUDA kernel where each warp is assigned to a window ($\vec{x}$) of the transformed image (\emph{Warp-level SVM}). The computation of the dot product is divided between the threads in the warp. Once intra-warp threads have computed the partial results, these are communicated among threads using register shuffle instructions and then reduced. Algorithm \ref{svm-pseudo} shows the mapping of the Sliding Window and SVM inference to the massive parallel architecture.

We decided to use a warp-level approach to avoid the overhead of thread synchronization as warps have implicit hardware synchronization. This configuration allows the full utilization of the memory bandwidth as the memory accesses are coalesced.
\begin{algorithm}[h] \label{svm-pseudo}
 \SetAlgoLined
\SetKwData{Left}{left}\SetKwData{This}{this}\SetKwData{Up}{up}
\SetKwInOut{Input}{input}\SetKwInOut{Output}{output}
 
 \Input{Fa[Hb][Wb][S], Fb[Hb][Wb][S] N[Hn][Wn][S]}
 \Output{scores[Y][X] }
 \BlankLine
  parallel \For {y=0 \KwTo Y and x=0 \KwTo X}{
   SIMD parallel \For {lane=0 to WarpSize}{
   t = lane\;
   \For{i=0 \KwTo Hm}{
     \For{j=0 \KwTo Wm}{
       \While{$t \textless S$}{
         res += Fa[i+y][j+x][t] $\ast$ N[i][j][t]\;
         res += Fb[i+y][j+x][t] $\ast$ N[i][j][t]\;
         t = t + \emph{WarpSize}\;
       }
     }
   }
   res = \emph{SIMDreductionSum}(res)\;
   \If{$lane == 0$}{
     scores[y][x] = res\;
   }
 }
 }
 \caption{Massively parallel computation of the Sliding Window and the SVM inference. \emph{N} is the SVM trained model, \emph{Hn} and \emph{Wn} are the number of Block Histograms fitting in a window and $S$ is the histogram size. $Hb\Leftarrow H/8-1$; $Wb \Leftarrow W/8-1$.}
\end{algorithm}
\vspace{-0.5cm}
\section{Experiments \& Results} \label{experiments}
\vspace{-0.3cm}
In this section we present the obtained performance results for the algorithms and pipelines. All the experiments are run with an Intel i7-5930K processor, and both Nvidia GPUs: GTX 960 and a Tegra X1. We start by showing the whole pipeline results. We measure the performance of each of the algorithms by measuring the number of processed pixels per nanosecond ($px/ns$); we will also refer to it as \emph{Performance}.



\vspace{-0.4cm}
\subsection{Detection Pipelines performance} \label{pipeline-res}
\vspace{-0.3cm}

\begin{figure}
\CenterFloatBoxes
\begin{floatrow}
\ffigbox{%
  \includegraphics[scale=0.35]{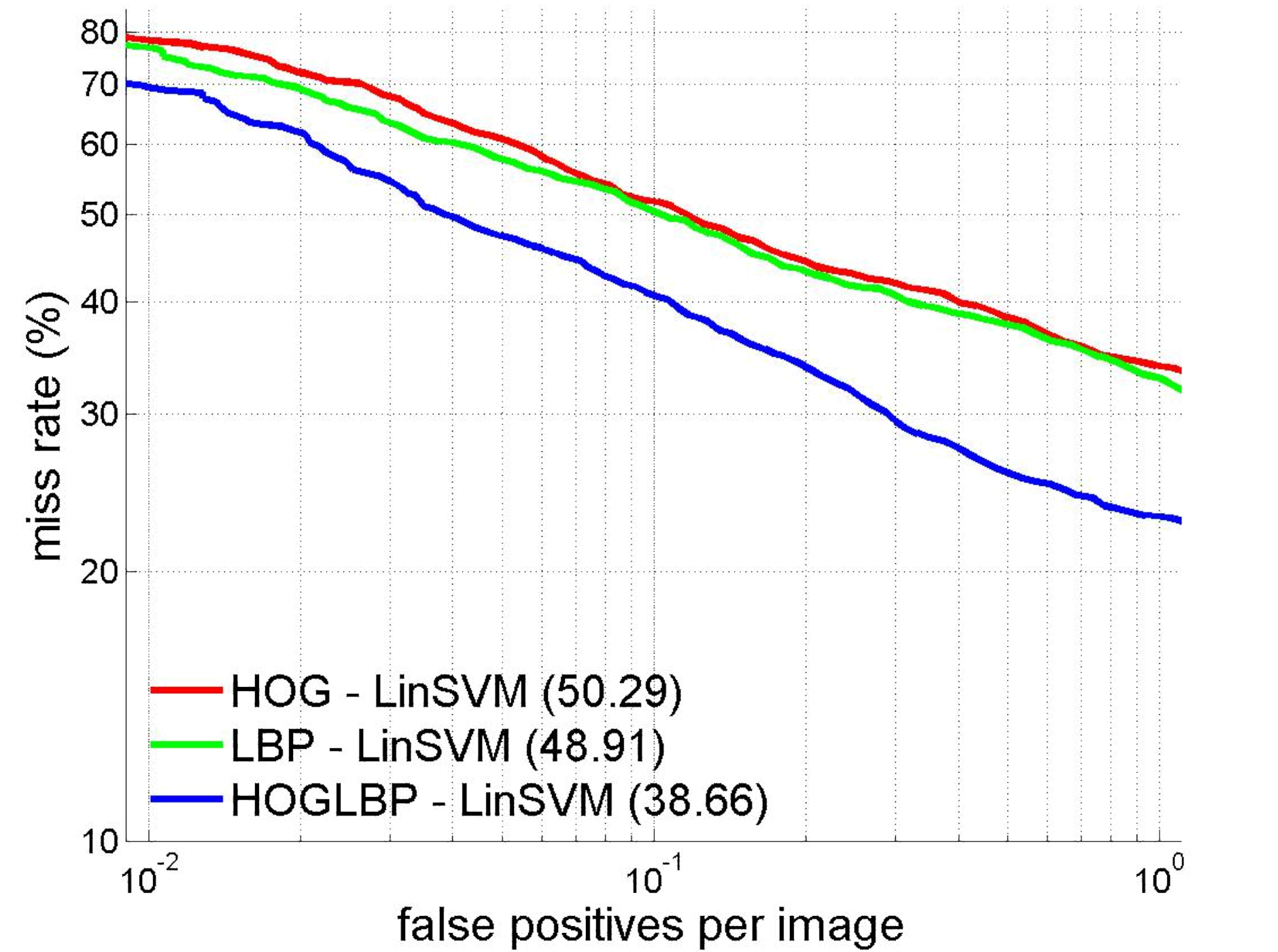}
}{%
  \caption{Legend values are the area below the curve. Must be minimized in reliable detectors
}
{
	\label{accuracy-res}
}
}
\ffigbox{%
  \includegraphics[scale=0.3]{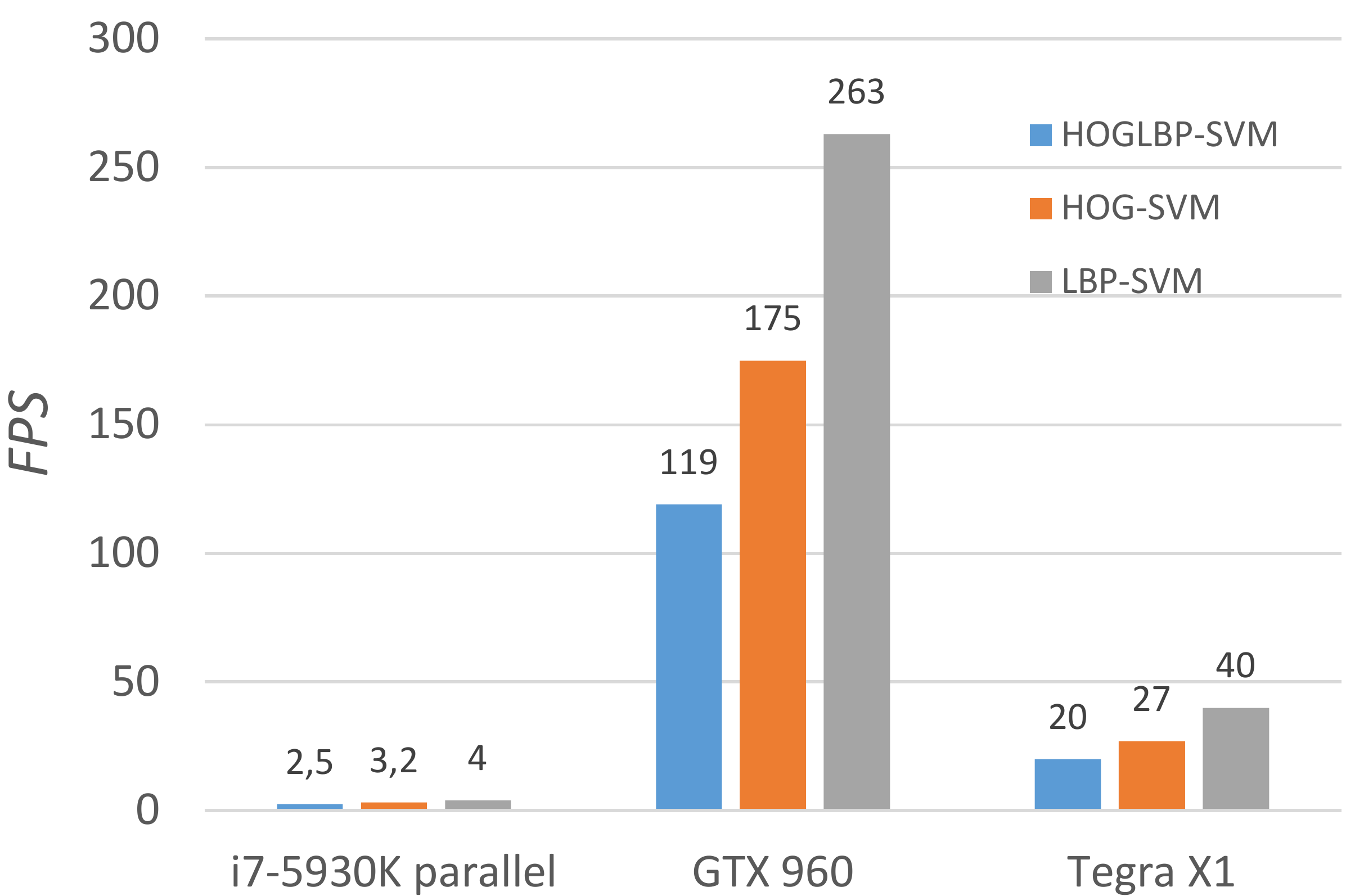}
}{%
  \caption{Performance of the detection pipelines measured in processed frames per second (FPS) in the different architectures.
}
{
	\label{perf-res}
}
}

\end{floatrow}
\end{figure}

To evaluate the $Performance$ we use a video sequence with an image size of  $1242\times 375$ pixels. Figure \ref{perf-res} presents the performance results of the LBP-SVM, HOG-SVM and HOGLBP-SVM pipelines, measured in processed frames per second (FPS). Results show the achieved FPS for the multithreaded CPU baseline application ~\cite{local-experts} and the GPU accelerated versions. Results prove that we have accomplished the objective of running the application in real-time under the low consumption ARM platform.

Figure \ref{accuracy-res} illustrates the miss rate depending on the false positive per image (FPPI). FPPI is the number of candidate windows wrongly classified as pedestrians, it can be understood as the tolerance of the system. As the FPPI increases, the miss rate decreases leading to a more tolerant system.

The system is able to achieve state of the art accuracy with the HOGLBP-SVM detection pipeline. The rest have slightly lower accuracy; nonetheless, they demand less computational power to achieve real-time performance which make them suitable for less powerful GPUs.
\vspace{-0.4cm}
\subsection{Detection Pipeline efficiency}
\vspace{-0.2cm}
Besides needing real time processing, the application demands low power consumption. To compare the power efficiency of the two GPUs we introduce a new metric: $FPS/Watt$. To assess the Watt consumption we use the Thermal Design Power (\emph{TDP}). \emph{TDP} is the amount of heat generated by the processor in typical use cases; the attribute is provided by the manufacturer company.

Table \ref{tperf-eff} shows the \emph{TDP} of each platform and the obtained efficiency. We expose the results of the HOGLBP-SVM pipeline as it is the pipeline that reaches state of the art accuracy. The Tegra X1 platform outperforms the CPU and GTX 960 implementations. It gives an increment of 200 points compared to the CPU application. It also doubles the efficiency of the GTX 960 desktop GPU.
\begin{table}[h]
  \begin{tabular}{l|c|c}
    HOGLBP-SVM & \emph{TDP} & $FPS/Watt$ \\
    \hline 
    i7-5930K & 140 & 0.01 \\
    GTX 960 & 120 & 0.99 \\
    Tegra X1 & 10 & 2 \\
  \end{tabular}
   \caption{\emph{TDP} of each platform and efficiency of the HOGLBP-SVM pipeline measured in $FPS/Watt$.}%
	\label{tperf-eff}
\end{table}
\vspace{-0.4cm}
\subsection{Feature extraction \& classification algorithms} \label{algo-res}
\begin{figure}[h]
\CenterFloatBoxes
\begin{floatrow}
\ffigbox{%
	\includegraphics[scale=0.4]{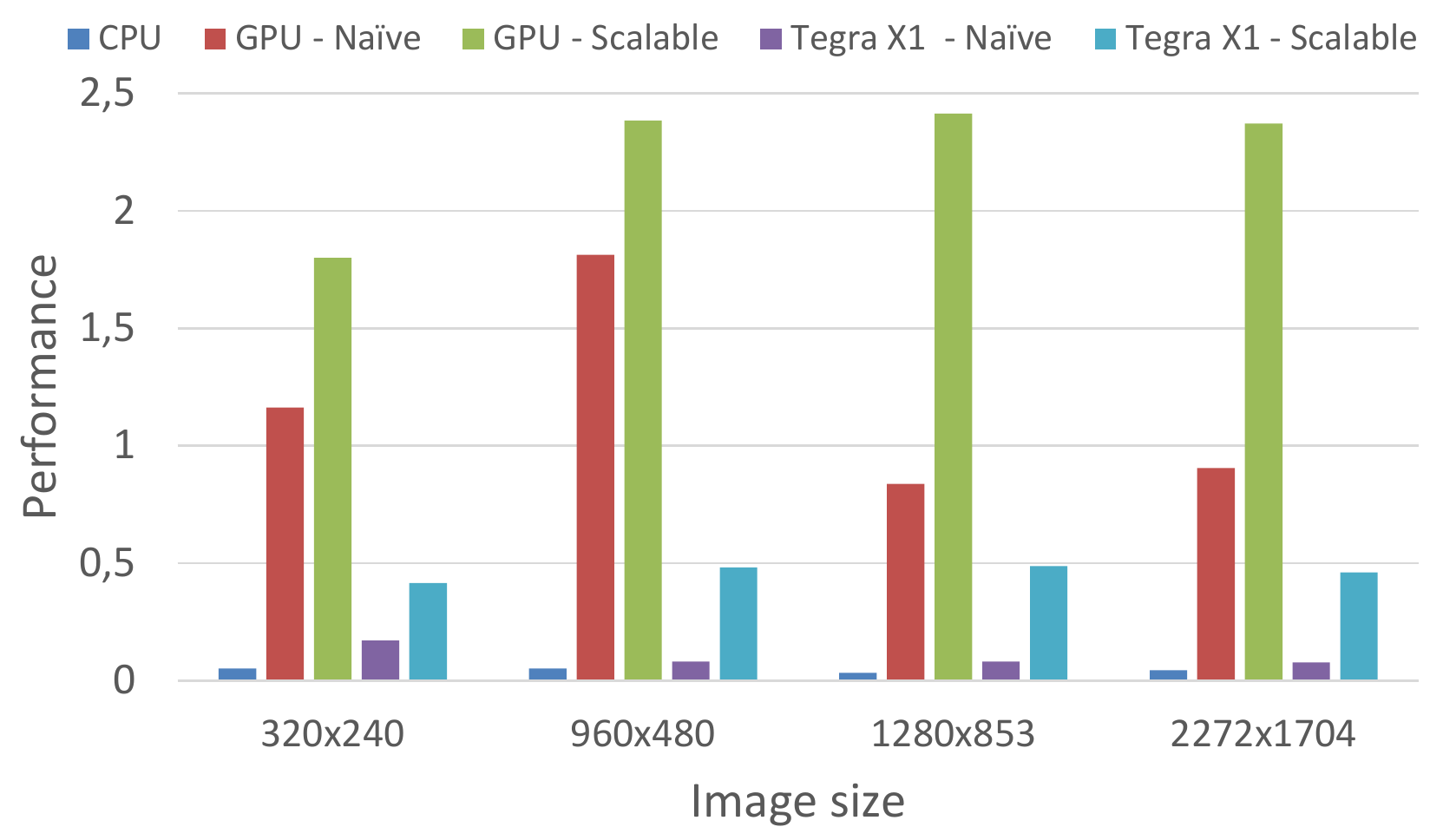}
}{%
  \caption{Performance of the LBP algorithm.}
{
	\label{perf-lbp}
}
}
\ffigbox{%
	\includegraphics[scale=0.4]{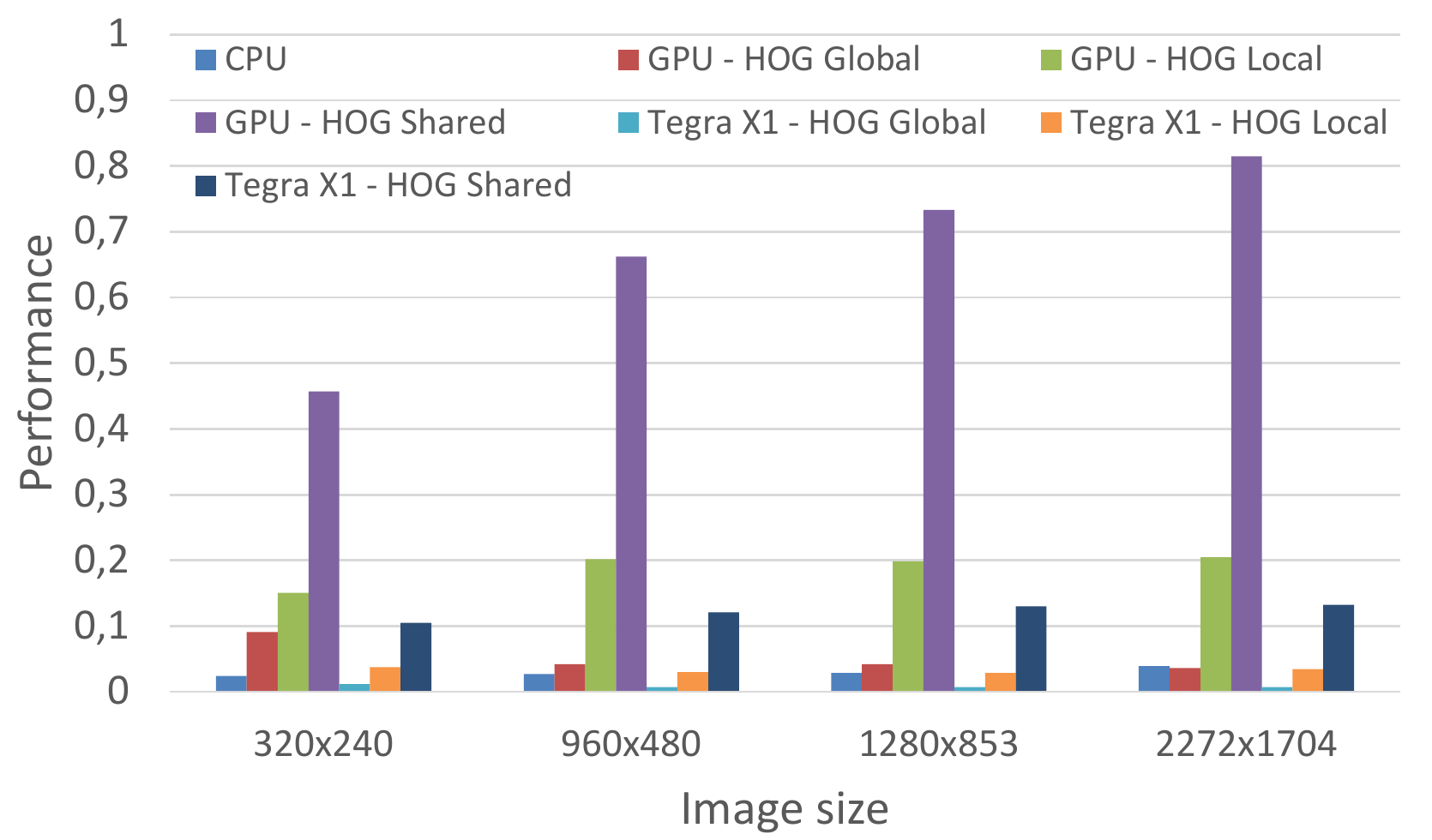}
}{%
  \caption{Performance of HOG algorithm.}
{
	\label{perf-hog}
}
}
\end{floatrow}
\end{figure}

\begin{figure}[h]
\centering
	\includegraphics[scale=0.4]{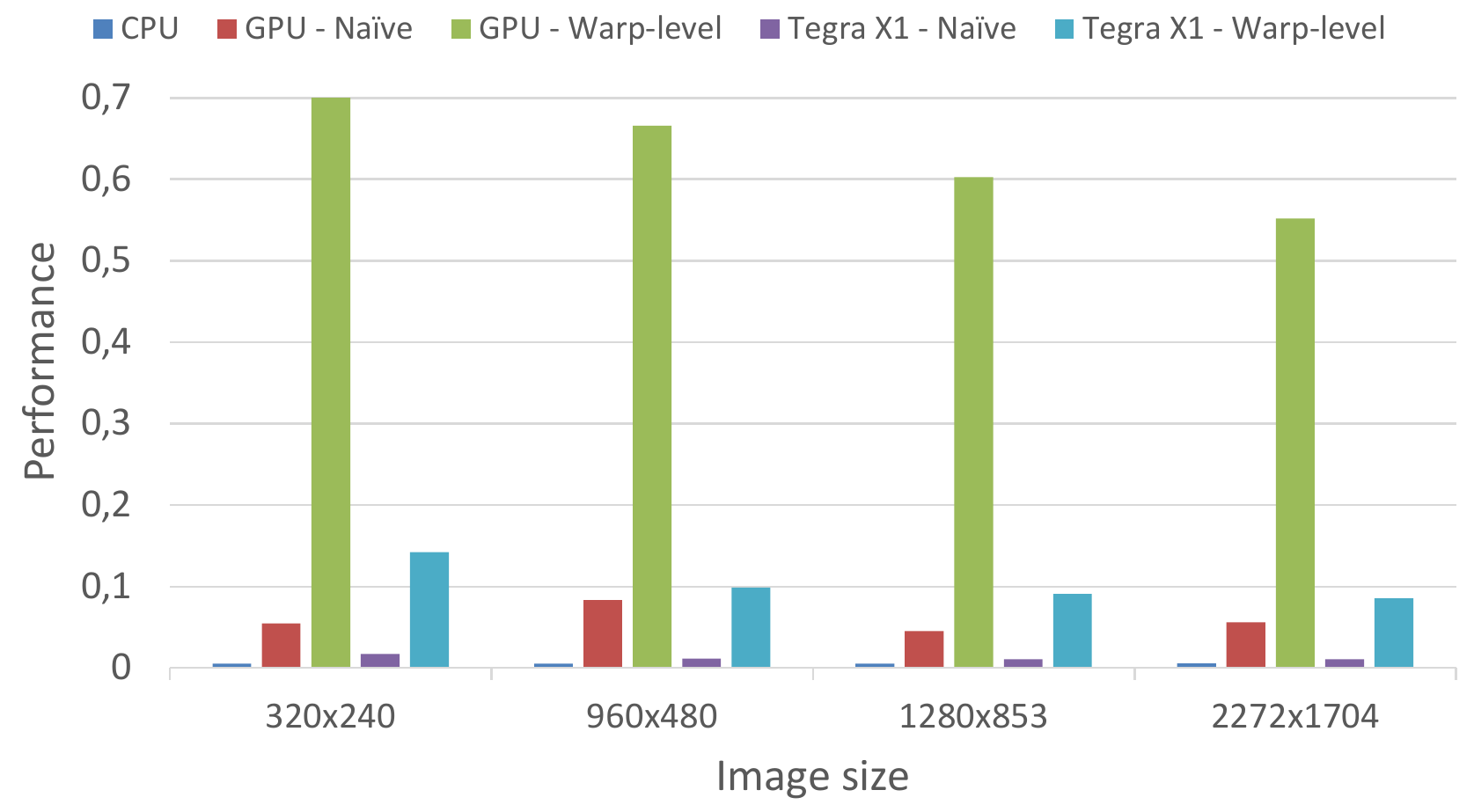}
  	\caption{Performance of the SW-SVM method.}
	\label{perf-svm}
\end{figure}

\vspace{-0.3cm}
In this section we present the performance results of the individual algorithms that form the three pipelines. Figures \ref{perf-lbp}, \ref{perf-hog}, \ref{perf-svm} show the obtained performance results measured in $px/ns$ with various image sizes.

For the LBP algorithm we experimented with the two developed designs (Na\"ive and Scalable schemes). As the image size increases, the Na\"{i}ve scheme suffers a decrease of performance caused by the poor memory management. However, the Scalable scheme performance is more regular because of the coalesced memory access (see Figure \ref{perf-lbp}). For the two biggest images it attains a 2.5x speed-up in both GPUs compared to the Na\"ive kernel. Although, for the smaller images the  boost is low; the non-coalesced memory access is attenuated by the high utilization of the L2 cache which hides the latency of the memory operations (Figure \ref{perf-lbp}).

Figure \ref{perf-hog} expose the attained results for the HOG algorithm. In this case, we analize the three developed solutions (HOG Global, HOG Local and HOG Shared). Results prove that the first massively parallel implementation was not competitive with a CPU version; for most of the experiments in Figure \ref{perf-hog}, the CPU outperforms the performance of the HOG Global kernel on both GPUs. However, if we consider the best implementation (HOG Shared) we can conclude that the performance the GPUs surpass the CPU; the GTX 960 has a speed-up of 4x and the Tegra X1 18x. The difference in speed-up between devices is caused by the lower memory bandwidth of the Tegra X1.
The HOG Shared kernel reduces the usage of the global memory as it stores the data in the shared memory leading to a less stressed memory in the Tegra ARM system.

The Pyramidal Sliding Window and Support Vector Machine is the most time consuming part of the pipeline it takes up to 55\% of the time to process an image. Figure \ref{perf-svm} shows the results for the Na\"ive and Warp-level approaches. The Na\"ive kernel suffers from non-coalesced memory access. It becomes a critical issue in the Tegra X1 as performance is equivalent to the CPU one. The Warp-level design the memory bandwidth is used at peak performance because of coalesced memory access. We obtain a boost of 10x compared to the Na\"ive kernel on the GTX960 card and 8x speed-up in the Tegra X1 system. Despite, the Warp-level kernel suffers a decrease of performance for the bigger images, in this cases, the data needs to be fetched from the global memory because it does not fit in the L2 cache. 
\vspace{-0.5cm}
\section{Conclusions} \label{conclusions}
\vspace{-0.3cm}
Our experiments confirm the importance of adapting the problem to the GPU architecture. Smart work distribution and thread collaboration are key factors to attain significant speed-ups compared to a high end CPU. The stated facts become even more critical when the development is done under a low consumption platform like the Tegra X1 processor. For the developed algorithms the $Performance/Watt$ of the Tegra X1 doubles the GTX 960. The evidence determines that the Tegra ARM platform is an energy efficient system able to challenge the desktop GPU performance when running massively parallel applications.

In this work we show a massively parallel implementation of a pedestrian detector that uses LBP and HOG as features and SVM for classification. Our implementation is able to achieve the real-time requirements on the autonomous driving platform, the Nvidia DrivePX.


\vspace{-0.4cm}
 \section{Acknowledgements}
\vspace{-0.3cm}
This research has been supported by the MICINN under contract number TIN2014-53234-C2-1-R. By the MEC under contract number TRA2014-57088-C2-1-R, the spanish DGT project SPIP2014-01352, and the Generalitat de Catalunya projects 2014-SGR-1506 and 2014-SGR–1562.
We thank Nvidia for the donation of the systems used in this work.
\vspace{-0.3cm}
%
\label{sect:bib}
\bibliographystyle{unsrt}
\bibliography{easychair}

\begin{thebibliography}{10}

\bibitem{geronimo-survey}
D.~Geronimo, A.~M. Lopez, A.~D. Sappa, and T.~Graf.
\newblock Survey of pedestrian detection for advanced driver assistance
  systems.
\newblock In {\em PAMI}, 2010.

\bibitem{gavrila-survey}
D.~M. Gavrila.
\newblock The visual analysis of human movement: A survey.
\newblock In {\em CVIU}, 1999.

\bibitem{caltech-survey}
Dollar, Wojek, Schiele, and Perona.
\newblock Pedestrian detection: An evaluation of the state of the art.
\newblock In {\em PAMI}, 2012.

\bibitem{local-experts}
J.~Marin, D.~Vazquez, A.~M. Lopez, J.~Amores, and B.~Leibe.
\newblock Random forests of local experts for pedestrian detection.
\newblock In {\em ICCV}, 2013.

\bibitem{fpga-hog}
M.~Hahnle, F.~Saxen, M.~Hisung, U.~Brunsmann, and K.~Doll.
\newblock {FPGA}-based real-time pedestrian detection on high-resolution
  images.
\newblock In {\em CVPR}, 2013.

\bibitem{ped-100-fps}
R.~Benenson, M.~Mathias, R.~Timofte, and L.~Van Gool.
\newblock Pedestrian detection at 100 frames per second.
\newblock In {\em CVPR}, 2012.

\bibitem{fastest-west}
P.~Dollar, Belongie, and Perona.
\newblock The fastest pedestrian detector in the west.
\newblock In {\em BMVC}, 2010.

\bibitem{sliding-win-gpu-wojek}
C.~Wojek, G.~Dorkó, A.~Schulz, and B.~Schiele.
\newblock Sliding-windows for rapid object class localization: A parallel
  technique.
\newblock In {\em Pattern Recognition}, 2008.

\bibitem{sliding-win-gpu-zhang}
Zhang and Nevatia.
\newblock Efficient scan-window based object detection using {GPGPU}.
\newblock In {\em CVPR}, 2008.

\bibitem{hog-svm-gpu}
V.~A. Prisacariu and I.~Reid.
\newblock fast{HOG} - a real-time {GPU} implementation of {HOG}.
\newblock In {\em Technical Report}, 2009.

\bibitem{hoglbp-wang}
X.~Wang, T.~X. Han, and S.~Yan.
\newblock An {HOG}-{LBP} human detector with partial occlusion handling.
\newblock In {\em ICCV}, 2009.

\bibitem{nms-laptev}
Laptev.
\newblock Improving object detection with boosted histograms.
\newblock In {\em Image and Vision Comp.}, 2009.

\bibitem{lbp-ojala}
T.~Ojala, M.~Pietikainen, and T.~Maenpaa.
\newblock Multiresolution gray-scale and rotation invariant texture
  classification with local binary patterns.
\newblock In {\em PAMI}, 2002.

\bibitem{hog-dalal}
N.~Dalal and B.~Triggs.
\newblock Histograms of oriented gradients for human detection.
\newblock In {\em CVPR}, 2005.

\bibitem{svm-vapnik}
C.~Cortes and V.~Vapnik.
\newblock Support-vector networks.
\newblock In {\em Machine learning}, 1995.

\end{thebibliography}





\end{document}